\documentclass[a4paper,twoside]{article}

\usepackage{epsfig}
\usepackage{subcaption}
\usepackage{calc}
\usepackage{amssymb}
\usepackage{amstext}
\usepackage{amsmath}
\usepackage{amsthm}
\usepackage{multicol}
\usepackage{pslatex}
\usepackage{apalike}
\usepackage{multirow}
\usepackage[bottom]{footmisc}
\usepackage{SCITEPRESS}     
\usepackage{todonotes}
\usepackage{graphicx}
\usepackage{algorithm2e}
\usepackage{interval}
\usepackage{booktabs}
\usepackage{mathtools}

\begin{document}

\title{Catch Me If You Can: Improving Adversaries in Cyber-Security With Q-Learning Algorithms}

\author{\authorname{Arti Bandhana\sup{1}\orcidAuthor{0000-0002-3711-3645}, Ondřej Lukáš\sup{1}\orcidAuthor{0000-0002-7922-8301}, Sebastian Garcia\sup{2}\orcidAuthor{0000-0001-6238-9910} and Tomáš Kroupa\sup{2}\orcidAuthor{0000-0003-1531-2990}}
\affiliation{Czech Technical University}
\email{bandhart@fel.cvut.cz, lukasond@fel.cvut.cz, sebastian.garcia@agents.fel.cvut.cz, tomas.kroupa@fel.cvut.cz}
}


\abstract{The ongoing rise in cyberattacks and the lack of skilled professionals in the cybersecurity domain to combat these attacks show the need for automated tools capable of detecting an attack with good performance. Attackers disguise their actions and launch attacks that consist of multiple actions, which are difficult to detect. Therefore, improving defensive tools requires their calibration against a well-trained attacker. In this work, we propose a model of an attacking agent and environment and evaluate its performance using basic Q-Learning, Naive Q-learning, and DoubleQ-Learning, all of which are variants of Q-Learning. The attacking agent is trained with the goal of exfiltrating data whereby all the hosts in the network have a non-zero detection probability. Results show that the DoubleQ-Learning agent has the best overall performance rate by successfully achieving the goal in $70\%$ of the interactions.}

\onecolumn \maketitle \normalsize \setcounter{footnote}{0} \vfill

\section{\uppercase{Introduction}}
\label{sec:introduction}

The risk of cyber attacks is constantly increasing. Attackers continue to become more sophisticated and manage to find new vulnerabilities to exploit, making the role of network defenders skewed and asymmetric. Most attack techniques involve little direct interaction between the attacker and the defender. In attacks such as ransomware~\cite{ransomware}, port scanning or cryptocurrency mining, the interaction can be as little as only one action from the attacker. In more complex attacks such as banking trojans or Advanced Persistent Threat (APT) attacks ~\cite{9213690}, the attacker has to perform a series of steps within the network or target device to be successful while remaining undetected. Such attacks are extremely difficult to detect, yet they are the most impactful. APT attacks are usually long-term, with many decisions typically taken by a human adapting their tactics and techniques to avoid detection and in most cases, the defense mechanisms are not versatile enough to adapt to the behavior of an attacker.

APT attackers can be modeled as agents who pursue their goals while interacting with an environment (target device or network). Most of these interactions are captured mainly by Game theory or Reinforcement Learning (RL) models with the intent of improving defenses in the network. Game-theoretic frameworks are used to provide solutions for optimal defenses (such as honeypot allocation) but RL models are mostly used to improve penetration testing attacks~\cite{Durkota2016,8301630}. LSTM network and Q-Learning techniques are also being applied to predict the attacker's action in APT data sets~\cite{dehghan_proapt_2022}. However, modeling realistic defenses inevitably requires learning almost optimal decisions for attackers. To the best of our knowledge, there are no studies about modeling APT attacker's behavior with the goal to \textit{improve} the decisions made by the attacker. Creating a realistic inference model for the attacker requires consideration of factors such as intent, capabilities, objectives, opportunities, and available resources for the attacker~\cite{Moskal2018,liu2005incentive}. Due to the complexity of these attributes, developing a general framework becomes challenging. To overcome these challenges, RL models are generally applied to train and solve an optimal policy from a defender's perspective; however, we are unaware of a RL model to optimize the actions of an APT attacker.

In this paper, we model both an APT attacker and a network environment to train RL agents that optimize the attack. The goal of the attacker is to exfiltrate data from a specific server inside a local network to a command and control (C\&C) server in the Internet. To find the optimal policy for the attacker, three off-policy RL algorithms are trained: Q-Learning, Naive Q-Learning, and DoubleQ-Learning.

Our results show that the DoubleQ-Learning-based attacker agent is able to exfiltrate data in almost $70\%$ of the interactions.
Furthermore, we show that the agent can learn how to plan and execute a multistage data exfiltration attack detected less than $40\%$ of the time. From a cybersecurity point of view, it means that a model of an attacker can be learned and improved, and therefore a better model of the defender could be learned in future research.

The main contributions of this paper are:
\begin{itemize}
    \item a novel model of a decision-making entity (APT attacker) in an adversarial environment;
    \item implementation of RL algorithms for an attacking agent in a custom environment; and
    \item an analysis of the impact of APT attacker models on the cybersecurity domain.
\end{itemize}

The paper is structured as follows. Section~\ref{sec:Motivation} provides the motivation and previous work. Section~\ref{sec:model-env} describes the RL environment. Section~\ref{sec:RL-algorithms} presents the RL algorithms; Section~\ref{sec:experiments-setup} presents the setup of the experiments; Section~\ref{sec:experiment-results} presents the results and discusses their impact. The conclusions and future work are contained in Section~\ref{sec:conclusions}.

\section{\uppercase{MOTIVATION \& RELATED WORK}}
\label{sec:Motivation}

There are two main sources of motivation for studying the behavioral models of attackers in APT attacks for local networks. First, improving defense mechanisms (algorithms, antivirus systems, etc.) based on the knowledge of past attacks highlights the need to better understand the characteristics of nearly optimal attack behaviors in realistic networks. Second, by creating and training RL models of the attacker's behavior, it is possible to optimize future defense mechanisms and the dynamic properties of such systems.

Game theory and RL~\cite{shiva2010game} have gained traction over the years in modeling attack and defense mechanisms in many domains, including network security. 
Network security problems are primarily complex and require rational decision-making. Game theory provides mathematical models of strategic interaction among multiple decision makers (players or agents) along with algorithms for finding solutions (equilibria) in such scenarios. The potential benefit of applying game theory to network security is the automation of the exhaustive threat detection process for network administrators. 
However, real-world cybersecurity models may have limitations with regard to the information observed by players. Typically, the defender's knowledge of the attacker’s strategy and decisions~ is  limited~\cite{patil2018applications}. This leads to games with partial observation or incomplete information, which are extremely difficult to scale to the required size of the problem. 

In the area of game theory for security, there has been promising research in honeypot technologies~\cite{9700616}. The authors designed an optimal approach for honeypot allocation by formulating a two-player zero-sum game between the defender and the attacker, which is played on top of an attack graph. The defender places honeypots on machines, while the attacker selects an attack path through the attack graph, which would lead to the target machine without being detected. In addition to solving an effective strategy for honeypot placement in the network, the authors also experiment with a diversity of honeypot configurations. Diversifying the honeypot configuration ensures that not all honeypots are discovered if one is compromised; however, this adds to the operational cost. To automate response to a cyber attack, \cite{hammar2020finding} investigate methods where strategies evolve without human intervention and do not require domain knowledge. The authors model the cyber interaction as a Markov game and use simulations of self-play where agents interact and update their strategies based on experience from previously played games.  

Another promising research direction used Proximal Policy Optimization (PPO) with self-play to solve a stochastic (Markov) two-player game with sequential moves between defender and attacker~\cite{du2022learning}. The game is played on top of an attack graph, and the authors show that the performance of a PPO policy is better than that of a heuristic policy. The initial results are promising, but the setting used by the authors is limited to the attack graph with five nodes and four edges. By contrast, our work deals only with a single-agent environment.

Attack graphs are helpful, as they can predict the attacker's path depending on the vulnerabilities present in the network. At the same time, defenders can leverage attack graphs to find an effective defense strategy. In particular,~\cite{https://doi.org/10.48550/arxiv.2112.13175} provides defense solutions through edge blocking in an attack graph constructed in the active directory. Another stream of research focuses on the assistance of attacking tools for better penetration testing or cyber-training, for example, using Deep Q-Learning~\cite{niculae2020automating}. The authors compare Q-Learning, Extended Classifier Systems (XCS), and Deep Q-Networks (DQN) to find attacker strategies. To determine the best response for a suspicious user on the network, \cite{7423125} compares the variations of Q-Learning with a stochastic game.

\section{ENVIRONMENT MODEL}
\label{sec:model-env}
Q-Learning is one of the most widely applied model-free off-policy RL algorithms ~\cite{8836506}. It allows agents to learn in domains with Markovian properties and thus can be modeled as a Markov Decision Process (MDP). 
Sufficient exploration of the environment is done with a $\epsilon$-greedy policy. An $\epsilon$-greedy method chooses a uniformly random action with probability $\epsilon$ and greedy action with probability $1-\epsilon$. 
The hyperparameter $\epsilon$ is chosen to balance exploration and exploitation, intending to maximize the cumulative reward.

An MDP is used as the underlying model ~\cite{sutton2018reinforcement} as the focus is on training a single attacking agent. Such an approach results in the defender being part of the environment. In real-life scenarios, successful detection requires several steps, from placement of the defensive measures, detecting and generating alerts, to evaluating and addressing threats. In this work, the defender is modeled as a stochastic and global part of the environment.

\subsection{Network} 
\begin{figure}[!ht]
\centering
   {\epsfig{file = 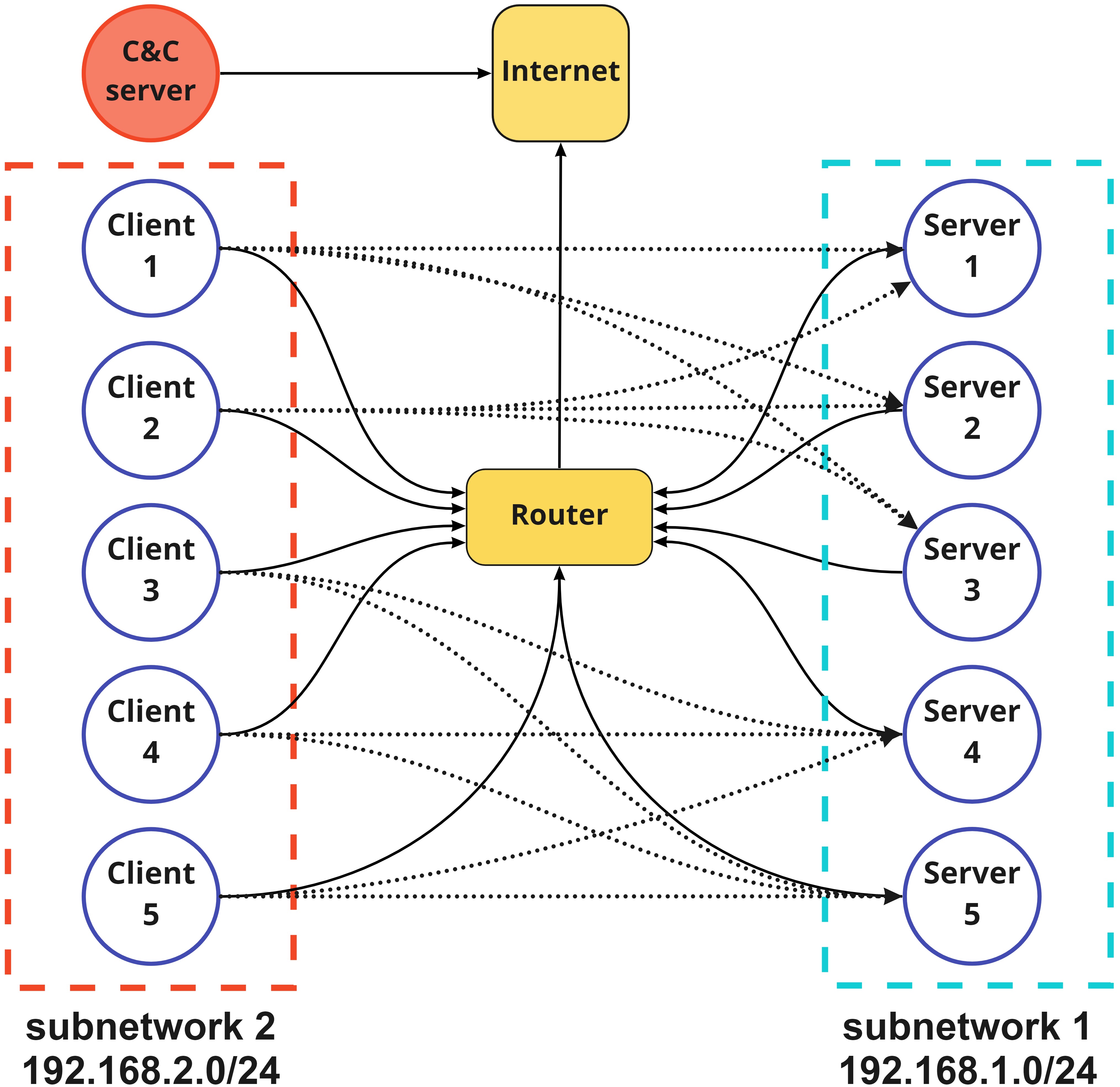, width = 5.5cm}}
  \caption{Network topology with two local subnetworks and a C\&C server on the Internet. The solid black lines represent direct network connectivity (such as Ethernet cables). The dotted lines represent logical connections from clients to servers as allowed by the firewall. In the non-randomized experiments, the attacker starts in Client 1. In the experiments with a randomized start, the attacker starts in one of the clients in subnetwork 2. The IP address of the C\&C server is always known to the attacker.}
  \label{fig:network}
 \end{figure}

The computer network used for the definition of the environment represents a small organization with five clients, five servers, and a router that provides Internet; see figure~\ref{fig:network}. Each host in the network has Internet access. The router is also a firewall that controls which clients from subnetwork 2 can access the servers in subnetwork 1 (corresponding to dotted lines in figure~\ref{fig:network}). Computers can connect to each other if they are in the same subnetwork. 

In the environment, we assume that the attacker has already gained access to one of the clients on the network. Additionally, the attacker knows the address of an external C\&C server on the Internet. The attacker's goal is to find and exfiltrate data located in one of the servers in subnetwork 1.


\subsection{Defender}
The defender in our model is an entity present in all clients/servers simultaneously and it has assigned a probability of detecting the attacker's action. Once the attacker is detected, the episode ends and the environment is reset to the initial state. This is represented by a terminal state in the environment. Given that the defender has full network visibility, there is a probability of detection for each action on all clients and servers. 

\subsection{Attacker}
Attackers usually do not have information about the network and so they must compensate for lack of knowledge by learning through trial and error. We simulate an attacker who has already gained a foothold in subnetwork 1 (figure~\ref{fig:network}) according to our assumption. This holds for a real-world scenario, as the initial breach can be done in various ways since there are many connected devices on the network, and preventing the initial breach in some ways is extremely hard. Therefore, modeling the attacker entry in our current setup is ignored. 
The attacker's objective is to find the optimal path to a server in subnetwork 2 containing sensitive data, find and exfiltrate this data, and make it accessible on the web. The available actions are the minimal actions required to complete the goal: find hosts, find services, get access, find data, and exfiltrate. The attacker was modeled as a rational attacker behaving optimally. 

\subsection{States}\label{subsec:states}
A state is an abstract representation of the environment from the attacker's perspective. It contains several assets the attacker can use or has discovered with previous actions. Therefore, the state of the environment changes based on the actions of the attacker and the current state. The probabilities $p(a|s)$ represent the probability of success of the attacker's action $a$ in a state $s$ and $p(detection|s,a)$ represents the probability of detection given the action $a$ played in the state $s$. These probabilities (table \ref{tab:action_prob}) of success and detection were set based on the expert evaluations of penetration test professionals, where knowledge of the domain was compared and matched with the evaluation of various detection tools for malicious behavior discovery shown in~\cite{9566808}.

\begin{table}[ht]
\caption{Probability of success and detection for each action executed by the attacker in the network.}
\label{tab:action_prob}
\begin{tabular}{lll}
\toprule
{Action} &
  \begin{tabular}[l]{@{}l@{}}Success \\ probability\end{tabular} &
  \begin{tabular}[l]{@{}l@{}}Detection \\ probability\end{tabular}\\
    \midrule
ScanNetwork &
   $0.9$ &
  $0.2$ \\
  
FindServices &
  $0.9$ &
  $0.3$ \\
   
ExecudeCode InService &
  $0.7$ &
  $0.4$ \\
    
FindData &
   $0.8$ &
  $0.1$ \\
ExfiltrateData &
 $0.8$ &
  $0.1$ \\
  
  \bottomrule
\end{tabular}%
\end{table}

Success probabilities are based on known tools and techniques. While network issues are the cause of the failure of most actions, in the case of \textit{ExecuteCodeInService} other problems such as service versions and exploits quality have to be taken into account. Detection probabilities consider the false positives found in real networks with benign traffic by a human player. Some actions, such as \textit{ScanNetwork} with ARP scan, are highly successful and barely recognized (off-the-shelf state-of-the-art IDS can not detect it~\cite{snort-arp}). Often, even if these scans are detected, such alerts are dismissed for the sake of limiting the False positives. The same applies to \textit{FindData} which is performed locally and thus nearly undetectable and \textit{ExfiltrateData} which when done correctly, is known to be extremely hard to distinguish from benign traffic.

\noindent At each time step, the following information is part of the state:
\begin{itemize}
    \item set of networks the attacker has discovered;
    \item set of hosts the attacker has discovered;
    \item set of hosts that the attacker has control of;
    \item set of services the attacker has discovered in each host; and
    \item set of data the attacker has discovered in a host.
\end{itemize}

Having states consisting of assets, we can follow the well-known STRIPS representation originally designed for planning~\cite{STRIPS}. STRIPS describes transitions in a system as operators, which are applicable if \textit{preconditions} are met. Originally, the effects of \textit{add} and \textit{delete} can be specified for each operator. However, in our approach, we completely omit the delete effect, which results in a \textit{relaxed} problem representation~\cite{STRIPS_relax}. Problem relaxation is a commonly used method in a variety of AI areas. Such an approach simplifies the problem of traversing the state space.

\subsection{Actions}
\label{subsec:actions}
The attacker's actions follow the subset of techniques for adversary behavior listed in Mitre ATT\&CK\footnote{https://attack.mitre.org/}. As we are only representing one type of goal in this model, data exfiltration, only the subset of Mitre actions related to data exfiltration are used:
\begin{enumerate}
   \item active scanning:
     \begin{enumerate}
     \item find computers in the network
     \item find services run on the hosts in the network
     \item find data in the computer 
     \end{enumerate}
    \item attack service to execute code; and
    \item exfiltrate data to the Internet.
    \end{enumerate}

The attacker in our model follows a five-step action as represented in Table \ref{tab:action_description} to reach its goal.

\begin{table}[ht]
\caption{List of actions and their effect on the network leading to a change in state}
\label{tab:action_description}
\resizebox{0.45\textwidth}{!}{%
\begin{tabular}{llll}
\toprule
{Action} &
  {Description} &
  {Preconditions} &
  {Effects} \\ \midrule
ScanNetwork &
  \begin{tabular}[c]{@{}l@{}}Scans complete\\  range of given network\end{tabular} &
  network + mask &
  extends 'known\_hosts' \\ 
  \midrule
FindServices &
  \begin{tabular}[c]{@{}l@{}}Scans given host\\ for running services\end{tabular} &
  host IP &
  \begin{tabular}[c]{@{}l@{}}extends 'known\_services'\\ with host:service pairs\end{tabular} \\ 
  \midrule
\begin{tabular}[c]{@{}l@{}}ExecudeCode\\ InService\end{tabular} &
  \begin{tabular}[c]{@{}l@{}}Runs exploit in service \\ to gain control of a host\end{tabular} &
  host:service &
  extends 'controlled\_hosts' \\ 
  \midrule
FindData &
  \begin{tabular}[c]{@{}l@{}}Runs code to discover \\ data in a controlled host\end{tabular} &
  hostIP &
  \begin{tabular}[c]{@{}l@{}}extends 'known\_data' with \\ host:data pairs\end{tabular} \\ 
  \midrule
ExfiltrateData &
  \begin{tabular}[c]{@{}l@{}}Moves data from one \\ controlled host to another\end{tabular} &
  host:data:host &
  \begin{tabular}[c]{@{}l@{}}extends 'known\_data' \\ with 'target:data'\end{tabular} \\ 
  \bottomrule
\end{tabular}%
}
\end{table}

\subsection{Rewards}\label{subsec:rewards}
The reward is an incentive that the agent receives with respect to the state action pair. In our model, the reward of the agent is constructed as:
$-1$ for every action taken,$-50$ if the action is detected, and $+100$ if the goal state is reached.

The small negative reward per action is intended to motivate the agent to find the shortest path to the goal. The $+100$ reward for the achievement of the goal allows the attacker to take actions with a higher expected detection probability if they lead to a higher expected reward.

\subsection{Implementation}

The representation of a state, as described in section~\ref{subsec:states}, allows the modification of the environment \textit{without} the need to retrain the agent from zero. This differentiates our environment model and offers a higher degree of modularity for various cybersecurity scenarios. Instead of allocating the complete Q-table prior to training, our agents create the Q-values dynamically, saving both memory and time during training. 

\section{LEARNING AGENTS} \label{agents}
\label{sec:RL-algorithms}

To train and evaluate the attacker’s performance, we use Q-Learning~\cite{8836506} and its variants: Naive Q-Learning and Double Q-Learning. Q-Learning is a reinforcement learning algorithm that approximates the optimal state-action value function independently of the policy being followed. It is an off-policy algorithm that separates learning from the current acting policy by updating the Q-value $Q(s, a)$, which is an indication of how good a state-action pair is. The equation for the Q-value update is:

\begin{equation}\label{eq1}
    Q(s,a) \coloneqq Q(s,a) + \alpha(R_{t+1} + \gamma V^{t}(s')),
\end{equation}

\noindent where $\alpha \in [0,1]$ is the learning rate and $\gamma \in [0,1]$ is the discount factor that captures the concept of depreciation. A value closer to $0$ means that the current reward is preferred over future rewards.

In Naive Q-Learning, the learning rate is partially allocated to the previous result to combine the knowledge of the past history during learning, the actual immediate reward in the current iteration, and the expected future reward~\cite{7423125}. This leads to the following variation of equation (\ref{eq1}):

\begin{equation}\label{eq2}
    Q(s,a) \coloneqq \alpha Q(s,a) + (1 - \alpha)(R_{t+1} + \gamma V^{t}(s'))
\end{equation}

Double Q-Learning~\cite{hasselt2010double,jang2019q} proposes learning two Q-functions instead of one. Each Q-function gets the update from the other for the next state. These two Q-functions are an unbiased estimate of the value of the action. The action selection is then performed by averaging or adding the two Q values for each action and then performing  $\epsilon$-greedy action selection  with the resulting Q values. In this paper, action selection is performed by adding the two Q values before performing the $\epsilon$-greedy. 

\begin{equation}\label{eq3}
    Q^{A} (s,a) \coloneqq Q^{A}(s,a)+\alpha(R +\gamma Q^{B}(s',a')-Q^{A}(s,a))
\end{equation}

\begin{equation}\label{eq4}
    Q^{B} (s,a) \coloneqq Q^{B}(s,a)+\alpha(R +\gamma Q^{A}(s',a')-Q^{B}(s,a))
\end{equation}

The other two learning agents also use $\epsilon$-greedy as the action selection criteria in accordance with the original papers.

\section{EXPERIMENT SETUP}
\label{sec:experiments-setup}

\textit{Three different scenarios} were used to train the learning agents: specific attacker position, random attacker position, and random target server to attack.

In the first scenario, the attacker is placed on client~1 in subnetwork 2 (figure~\ref{fig:network}). We define a client as an official device on the network used for work and a server as a device that holds data and offers services accessed by the clients. The attacker's goal is to reach the target server, which is specified as server 3 in subnetwork 1; exfiltrate the data from the target server to the C\&C server outside the local network. 

There are five clients in subnetwork 2, and in reality, any connected device within the network is susceptible to an attack; therefore, for the second scenario, we randomly assign the starting position of the attacker. This was done to compare the performance of the learning agents and see how they adapt to randomness in the starting position. In addition to randomizing the starting position, we also randomized the target server for data exfiltration; which was our third scenario. 

For successful achievement of the goal state, at least $5$ successful actions had to be performed in all $3$ scenarios; however, if the agent exceeds the limit of $25$ actions per episode, the interaction is terminated.


The defender in all $3$ scenarios is an entity with unlimited visibility and is present in all hosts, that is, every action can be detected with a predefined probability. Additionally, we assume that all services running on the hosts are exploitable and that a connection to the Internet is available on all hosts.

The learning parameter for each algorithm is presented in Table~\ref{tab:A1}. Experiments start with a random attacker which randomly picks an action. The Q-Learning agent and the DoubleQ-Learning agent were trained on a learning rate of 0.3, while the Naive Q-Learning agent was trained on a learning rate of 0.8. The action selection parameter controlled by epsilon was kept at 0.2 for all the agents; however, Double Q-Learning used a linearly decaying $\epsilon$ from 0.2 to 0.05.

In all experiments, we measured the win rate, the detection rate, and the mean return of the episodes. The win rate represents the percentage of interactions that were successful for the attacker, which is the number of times the attacker was able to reach the goal state and exfilter the data in $10\,000$ episodes. The detection rate represents the percentage of interactions that were detected and resulted in the attacker receiving a reward of $-50$. 

\begin{table}[ht]
\caption{Training parameters: Q-Learning and DoubleQ-Learning agents were trained in $10\,000$ episodes, while NaiveQ-Learning was trained in $5\,000$ episodes. The discount factor $\gamma$ was kept at 0.9 for all learning agents. }
\label{tab:A1} \centering
\begin{tabular}{lcccc}
\toprule
Algorithm                                                    & $\alpha$   & $\epsilon$ & $\gamma$  & \begin{tabular}[c]{@{}l@{}}No. of  \\ episodes\end{tabular} \\ 
\midrule
Random  & -   &   - & - & - \\ 
Q-Learning  & 0.3 & 0.2 &0.9 & 10\,000 \\ 
Naive Q-Learning  & 0.8 & 0.2 &0.9 & 5\,000 \\ 
Double Q-Learning & 0.3 & 0.2 &0.9 & 10\,000 \\
\bottomrule
\end{tabular}
\end{table}

\section{EXPERIMENTAL RESULTS}

\label{sec:experiment-results}
The following results were obtained in the first scenario of the experiment, when the attacker's position was specified in the network. Table ~\ref{tab:1} summarizes the performance of the different learning agents. The random attacker, without any knowledge of the network and without any strategy, has a detection rate of $99.58\%$, while the DoubleQ-Learning attacker had a detection rate of $33\%$. The Q-learning and Naive Q-Learning agents have similar detection rates. 

\begin{table}[ht]
\caption{Performance comparison of the learning agents with a fixed attacker starting in \textit{client1}. Q-Learning and Double Q-Learning were trained with $10\,000$ episodes, while Naive Q-Learning was trained with $5\,000$ episodes.}
\label{tab:1}
\resizebox{0.45\textwidth}{!}{%
\begin{tabular}{lccc}
\toprule
    Algorithm &
  \begin{tabular}[c]{@{}l@{}}Winning \\ rate (\%)\end{tabular} &
  \begin{tabular}[c]{@{}l@{}}Detection\\ rate (\%)\end{tabular} &
  \begin{tabular}[c]{@{}l@{}}Mean\\ return\end{tabular} \\ 
\midrule
Random   & 0.48  & 99.58 & 53.03 \\ 
Q-Learning   & 66.4  & 40.4  & 43.94 \\ 
Naive Q-Learning  & 66.91 & 40.19 & 43.94 \\ 
Double Q-Learning & 74.0  & 33.0  & 54.61 \\
\bottomrule
\end{tabular}%
}
\end{table}

\begin{table}[ht]
 \caption{Comparison of performance for the learning agents in a scenario with  randomized attacker's starting randomized. Q-Learning and Double Q-Learning were trained with 10\,000 episodes, while Naive Q-Learning was trained with 5\,000 episodes.}\label{tab:2} 
 \resizebox{0.45\textwidth}{!}{%
 \begin{tabular}{lccc}
 \toprule
 Algorithm &
   \begin{tabular}[c]{@{}l@{}}Winning \\ rate (\%)\end{tabular} &
   \begin{tabular}[c]{@{}l@{}}Detection\\ rate (\%)\end{tabular} &
   \begin{tabular}[c]{@{}l@{}}Mean\\ return\end{tabular} \\ 
   \midrule
 Random & 0.34  & 99.48 & -54.04 \\ 
Q-Learning   & 65.4  & 39.27 & 41.97 \\ 
Naive Q-Learning  & 54.27 & 50.78 & 25.59 \\ 
Double Q-Learning & 68.9  & 36.8  & 47.58 \\ 
\bottomrule
 \end{tabular}%
}
 \end{table}

\begin{table}[ht]
 \caption{Comparison of learning agents in a scenario where the attacker's starting point and target server were randomized. All algorithms were trained on 10\,000 episodes.}\label{tab:3} 
 \resizebox{0.45\textwidth}{!}{%
 \begin{tabular}{lccc}
 \toprule
 Algorithm &
   \begin{tabular}[c]{@{}l@{}}Winning \\ rate (\%)\end{tabular} &
   \begin{tabular}[c]{@{}l@{}}Detection\\ rate (\%)\end{tabular} &
   \begin{tabular}[c]{@{}l@{}}Mean\\ return\end{tabular} \\ 
   \midrule
    Q-Learning   & 53.3  & 53 & 23.45 \\ 
Naive Q-Learning  & 61.8 & 44.1 & 36.8 \\ 
Double Q-Learning & 64.9  & 41.7  & 41.2 \\ 
\bottomrule
 \end{tabular}%
}
 \end{table}

Randomizing the starting position decreases the win rate and increases the detection rate for all learning agents, as shown in table~\ref{tab:2}. The Naive Q-Learning agent had the greatest impact on performance due to the randomness of the starting position among all learning agents. The detection rate increased from $40.4\%$ to $50.78\%$

\begin{figure}[!ht]
\centering
   {\epsfig{file = 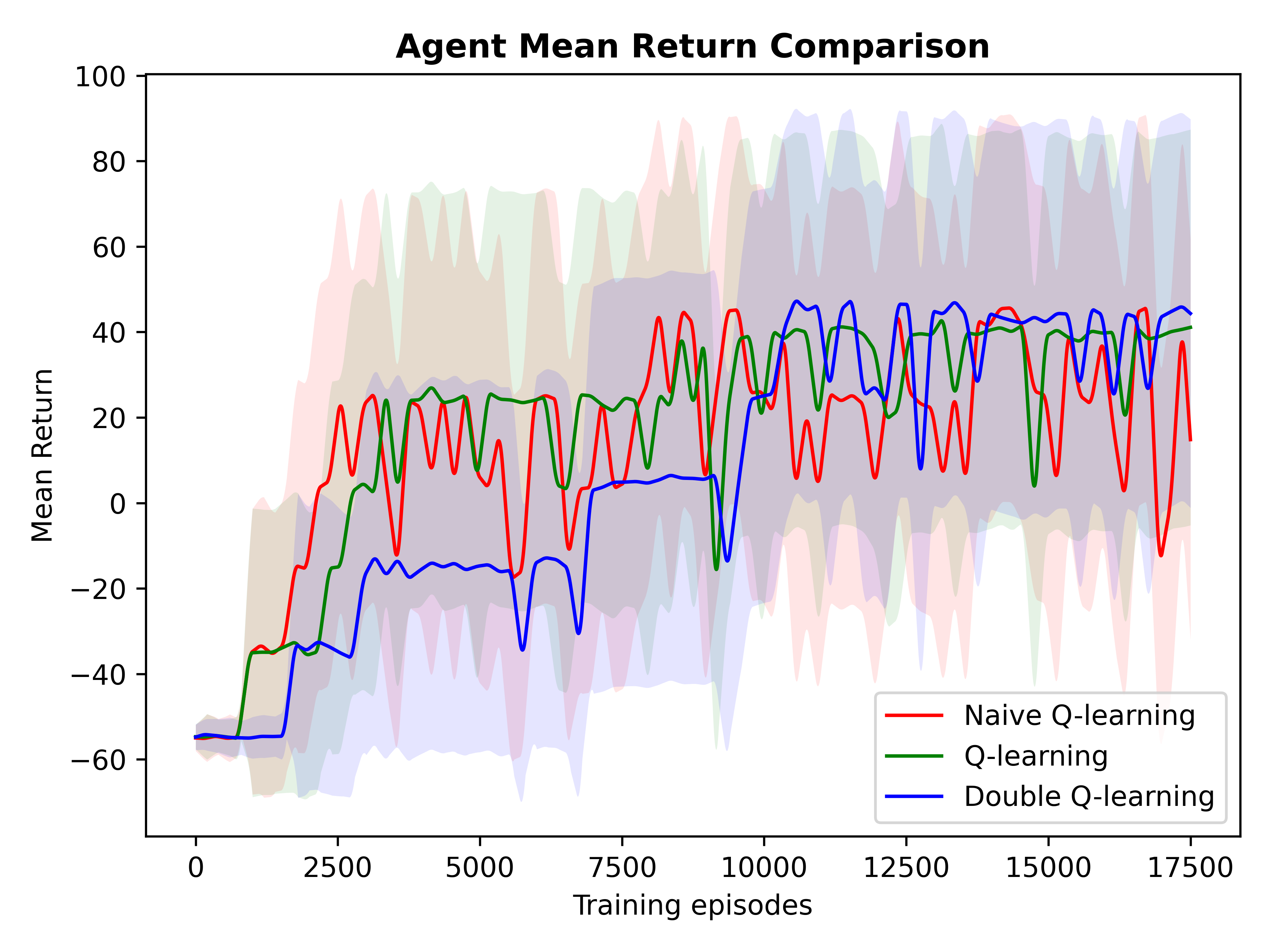, width = 0.45\textwidth}}
  \caption{Comparison of the mean cumulative reward of agents during the learning process in the scenario with defender and randomized starting position for the attacker.}
  \label{fig:learning_comparison_mean}
 \end{figure}

 \begin{figure}[!ht]
\centering
   {\epsfig{file = 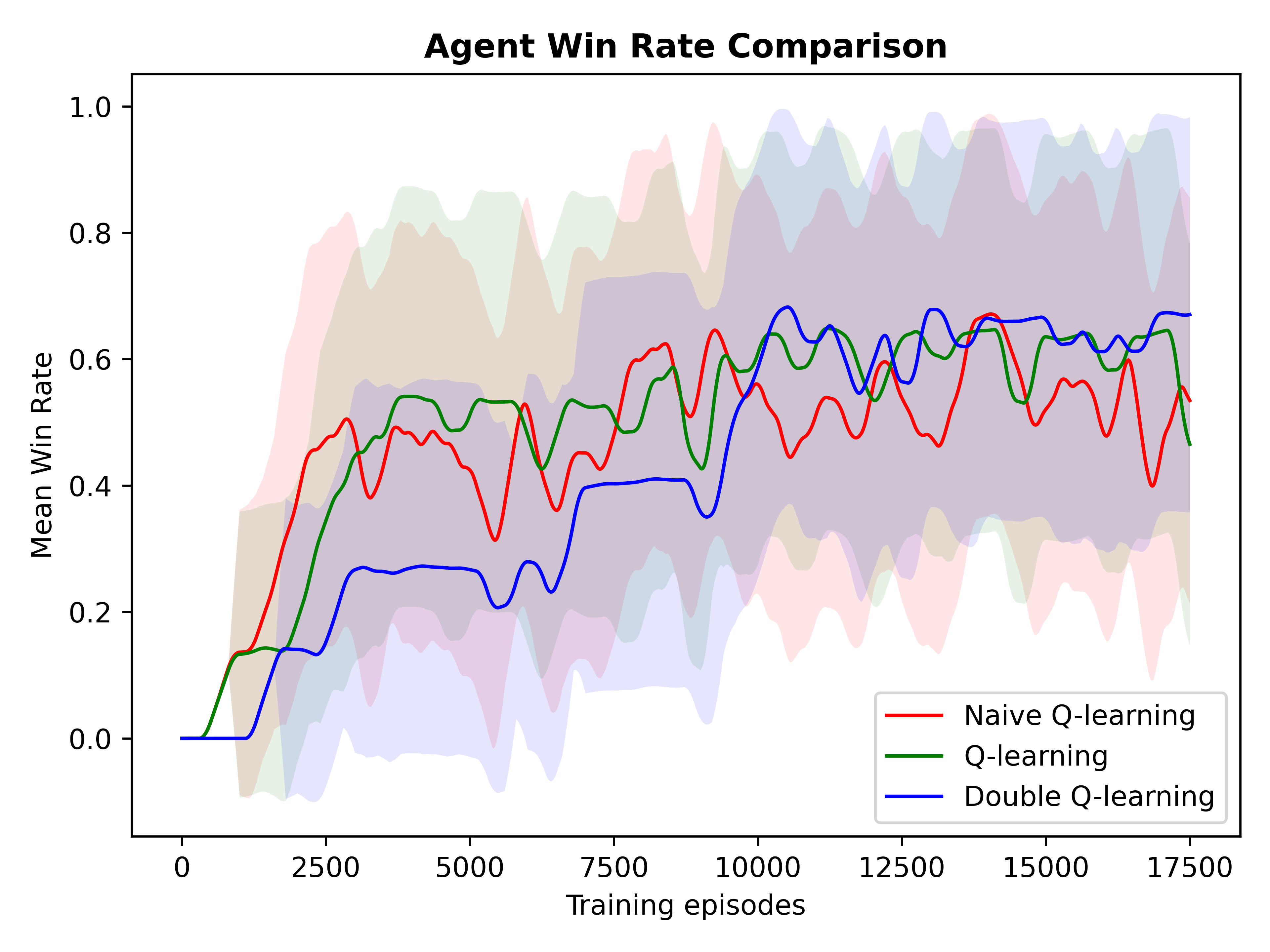, width = 0.45\textwidth}}
  \caption{Comparison of the winning rate of agents during the learning process in the scenario with defender and randomized starting position.}
  \label{fig:learning_comparison_wins}
 \end{figure}

\begin{figure}[!ht]
\centering
   {\epsfig{file = 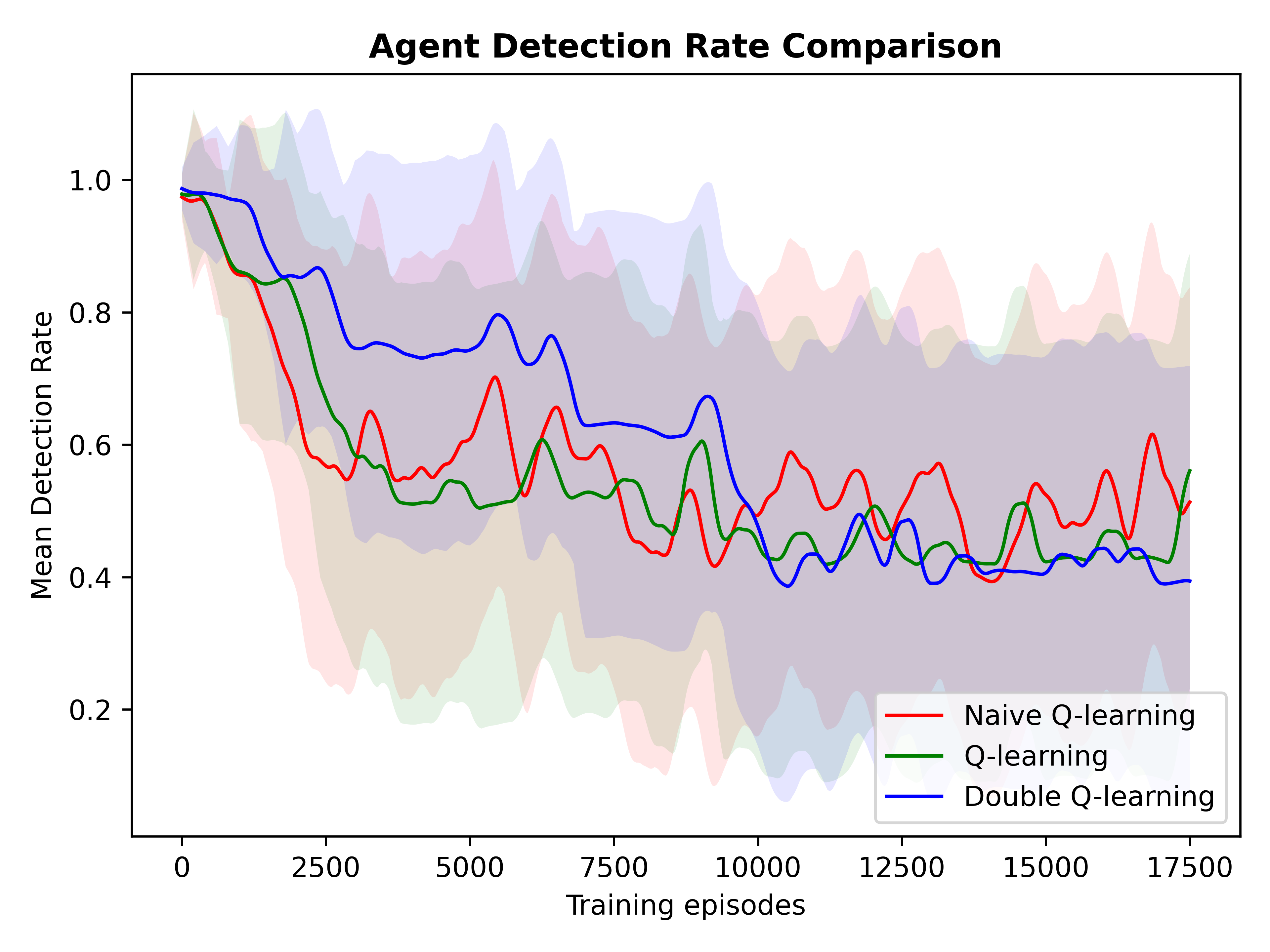, width = 0.45\textwidth}}
   \caption{Comparison of the detection rate of agents during the learning process in the scenario with defender and randomized starting position.}
  \label{fig:learning_comparison_detection}
\end{figure}

\subsection{Analysis of Results}\label{subsec:result_analysis}
We compared how agents with varying parameters learned a policy in a network with ten hosts in the presence of a defender with full visibility. The detection probability was nonzero for actions at all clients and servers. When comparing the win rate and the detection rate for all learning agents, it is clear that Double Q-Learning outperforms all other agents in all scenarios.
Two Q-functions are trained in such agent but from different episodes which makes the training more robust. A sum of the Q-functions is used during inference.
This avoids the overestimation bias of Q-Learning and leads to better training stability even in a noisy environment. The Q-Learning attacker and the Naive Q-Learning attacker have the same performance for the first scenario where the starting point was specified. This is due to the distribution of the learning rate according to equations (\ref{eq1}) and (\ref{eq2}). A learning rate of 0.8 was used for Naive Q-Learning, which in comparison with the Q-learning gives similar results as the learning rate of 0.2. However, the performance of Naive Q-Learning decreased when the starting position was randomized. This is attributed to the weighting of the update rule of the Q-value, as shown in equation (~\ref{eq2}). When considering negative rewards, the update affects the Q-value more than the standard Q-Learning due to the split update $\alpha$. Although this can be beneficial in the cases of high positive rewards, the results show that this approach lacks adaptability in the case of the stochastic environment.

 Figures~\ref{fig:learning_comparison_wins} and~\ref{fig:learning_comparison_detection} show that DoubleQ-Learning outperforms the other two agents in terms of winning and detection rates. The high variance of the mean returns, as shown in figure~\ref{fig:learning_comparison_mean} is the result of the stochastic environment and the reward distribution described in 
 section~\ref{subsec:rewards}. The graphs also show that even though DoubleQ-Learning performs badly in the beginning, over time as the number of episodes increases and state-action values are updated, it outperforms the other two learning agents. 
 In particular, even if the agent's policy is optimal, it cannot influence the detection and subsequent reward of $-50$. Therefore, the three agents share similar high variance in mean returns but differ significantly in metrics that focus on reaching the goal, in which the Double Q-Learning shows the most promising results.

 Despite using random exploration $\epsilon$ in the three agents based on Q-Learning, the results from the first and second scenarios show that the environment and the goal are non-trivial and unsolvable for agents performing purely random actions, which reached the goal in fewer than $1\%$ of the cases. For that reason, the Random Agent was excluded from the comparison in figures ~\ref{fig:learning_comparison_mean},~\ref{fig:learning_comparison_wins} and~\ref{fig:learning_comparison_detection} and in the \textit{third scenario}.

 The results of our experiments show that despite the defender having full visibility of the network, a rational attacker was still able to reach the target and exfiltrate data. From a security perspective, this indicates that the defensive tools in the network need to be improved so as to prevent the attacker's lateral movement in the system. 

\section{CONCLUSION}
\label{sec:conclusions}
  
In this paper, we propose a Q-Learning-based attacking agent capable of performing data exfiltration. 

Our results show that even though the three learning agents can find meaningful policies, Double Q-Learning outperforms the others and provides the most stable training. It reached the goal ~$70\%$ of the interactions while being undetected in $37\%$. This shows that despite a globally present defender, a rational attacker could still reach the target. 

The initial success and detection probabilities were set based on expert knowledge, however, our results clearly show that there is room for improvement in the detection capability of the defender. Having a high success probability for attacker action highlights the need for a robust defense mechanism that is capable of detecting any stealthy attacker.   
This provides a foundation for studying and improving attacker techniques to increase defense capability in the network.  

Currently, the method is limited to small or medium-sized networks. Although the interaction and world representation model can be easily extended to a more complex setup in size of the network and action space, the scalability and computational feasibility of such extensions have yet to be evaluated. 

Therefore, the natural direction for future research is to expand our approach towards larger environments, which will require subsequent scalability testing due to those complex setups. We also plan to incorporate other types of cyber attacks into Mitre taxonomy and model the defender as a rational entity with its own set of actions in the interaction. In addition, we plan to test the performance of our agent in a simulated environment.

Along with increasing the environmental complexity, the problem of more complex goals for the attacker is also in the pipeline resulting in the need for more reconnaissance from the agent. 

\section*{ACKNOWLEDGMENTS}
\label{sec:ack}

The authors acknowledge support from the Research Center for Informatics (CZ$.02.1.01\slash0.0\slash0.0\slash16\_019/0000765$) and Strategic Support for the Development of Security Research in the Czech Republic 2019--2025 (IMPAKT 1) program, by the Ministry of the Interior of the Czech Republic under No. VJ02010020 -- AI-Dojo: Multi-agent testbed for the research and testing of AI-driven cyber security technologies.

\bibliographystyle{apalike}
{\small
\bibliography{main}}
\end{document}